\documentclass[10pt,journal,compsoc]{ieeeojies}
\ifCLASSOPTIONcompsoc
  \usepackage[nocompress]{cite}
\else
  \usepackage{cite}
\fi
\usepackage{amsmath,amssymb,amsfonts}
\usepackage{algorithm2e}
\usepackage{graphicx}
\usepackage{textcomp}
\usepackage{lipsum}

\usepackage{amsthm}
\usepackage{makecell}

\DeclareMathOperator*{\argmin}{arg\,min}
\theoremstyle{definition}
\newtheorem{definition}{Definition}[section]
\usepackage{fancyhdr}

\usepackage{xcolor}
\definecolor{RoyalBlue}{cmyk}{1, 0.50, 0, 0}

\theoremstyle{remark}

\def\BibTeX{{\rm B\kern-.05em{\sc i\kern-.025em b}\kern-.08em
    T\kern-.1667em\lower.7ex\hbox{E}\kern-.125emX}}

\fancypagestyle{FirstPage}{
\lfoot{This work has been submitted to the IEEE for possible publication. Copyright may be transferred without notice, after which this version may no longer be accessible.}
}

\begin{document}
\title{Generalized Relevance Learning Grassmann Quantization}


\author{
        M.~Mohammadi,
        M. Babai~
        and~M.H.F.~Wilkinson,~\IEEEmembership{Senior Member,~IEEE}
\IEEEcompsocitemizethanks{\IEEEcompsocthanksitem M. Mohammadi is with the Faculty of Law, Department of Legal Methods, University of Groningen.\protect\\
E-mail: m.mohammadi@rug.nl
\IEEEcompsocthanksitem M.H.F. Wilkinson is with Bernoulli Institute, University of Groningen.\protect\\
Email: m.h.f.wilkinson@computer.org
\IEEEcompsocthanksitem M.Babai is with the Center for Information Technology (CIT), University of Groningen.\protect\\
Email: m.babai@rug.nl
}
        }

%
\begin{abstract}
Due to advancements in digital cameras, it is easy to gather multiple images (or videos) from an object under different conditions. Therefore, image-set classification has attracted more attention, and different solutions were proposed to model them. A popular way to model image sets is subspaces, which form a manifold called the Grassmann manifold. In this contribution, we extend the application of Generalized Relevance Learning Vector Quantization to deal with Grassmann manifold. 
The proposed model returns a set of prototype subspaces and a relevance vector. While prototypes model typical behaviours within classes, the relevance factors specify the most discriminative principal vectors (or images) for the classification task. They both provide insights into the model's decisions by highlighting influential images and pixels for predictions. Moreover, due to learning prototypes, the model complexity of the new method during inference is independent of dataset size, unlike previous works.
We applied it to several recognition tasks including handwritten digit recognition, face recognition, activity recognition, and object recognition.
Experiments demonstrate that it outperforms previous works with lower complexity and can successfully model the variation, such as handwritten style or lighting conditions. 
Moreover, the presence of relevances makes the model robust to the selection of subspaces' dimensionality. 
\end{abstract}



\maketitle
\thispagestyle{FirstPage}
\section{Introduction}
The presence of an increasing number of cameras in everyday life facilitates collecting multiple images (or videos) for a subject. These images can capture different forms of changes such as variations in illumination conditions, poses, facial expressions, etc. This can lead to building more robust models against possible changes. Therefore,  image-set classification has attracted more attention among researchers.

Linear subspaces have widely been used to model data points, and it has been shown that they form a non-linear manifold, called the Grassman manifold. Consequently, many techniques, developed for Euclidean spaces, are not applicable to this scenario \cite{huang2015projection}. We refer the reader for a broader application of Grassmann manifold to \cite{zhang2018grassmannian}.
Linear subspaces have also been extensively used to model image sets, 
and as a result, several methods \cite{yamaguchi1998face, fukui2005face, kim2007boosted,  kim2007discriminative, wang2008manifold, wang2009manifold, hamm2008grassmann, harandi2011graph, huang2015projection, wei2020prototype, sogi2020metric} were developed for dealing with subspaces and the Grassmann manifold. 
For example, methods presented in DCC\cite{kim2007discriminative}, MSM\cite{yamaguchi1998face}, CMSM\cite{fukui2005face}, and MMD\cite{wang2008manifold}, use canonical correlation/angle to evaluate the similarity between subspace and then use the Nearest Neighbor strategy for classification. While they use non-adaptive (dis-)similarity measures, others propose adaptive metrics to enhance separating different classes. 
BoMPA\cite{kim2007boosted} first tries to find important principal angles for the classification task and then puts more emphasis on them.
By generalizing the concept of canonical correlation, AMLS\cite{sogi2020metric} maximizes the separability between classes and the compactness within classes. 
PML\cite{huang2015projection} achieves a similar goal by learning a Mahalanobis-like metric. 
In contrast to them, GDA\cite{hamm2008grassmann} and GGDA\cite{harandi2011graph} use the kernel trick and embed points on the Grassmann manifold to high-dimensional Reproducing Kernel Hilbert Space (using Grassmann kernels) and then apply discriminant analysis. Unlike previous works, G(G)PLCR\cite{wei2020prototype} tries to directly model each classes 
and variations within them.

\begin{figure*}
    \centering
    \includegraphics[scale=0.9]{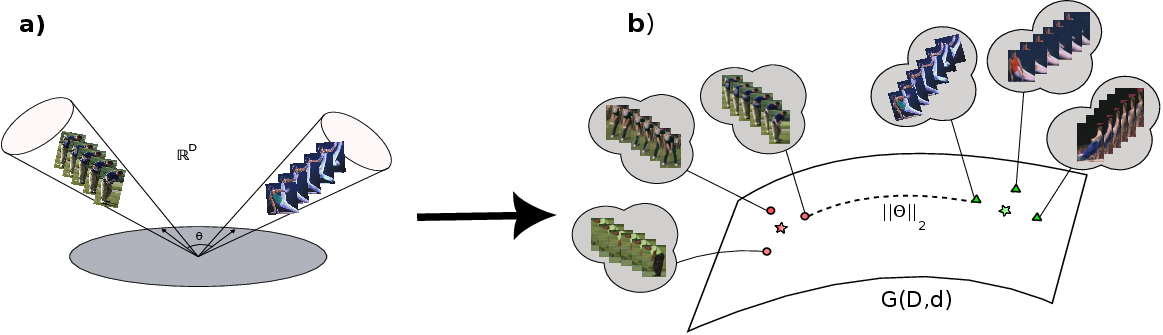}
    \caption{A conceptual demonstration of the proposed algorithm. a) image sets are modeled as d-dimensional subspaces. b) the set of all $d$-dimensional subspaces forms a specific type of Riemannian manifold, called Grassmann manifold. Thus, image sets can be represented as points on the manifold $G(D,d)$. The $l_2$ norm of principal angles defines the geodesic distance between any two points on the manifold. The proposed technique learns a set of prototypes in the data space (stars) capturing the essence of classes. Each color represents a different class (label).}
    \label{fig:grassmann_mdf}
\end{figure*}

The Generalized Learning Vector Quantization (GLVQ) classifiers is a family of models in which a set of prototype vectors models the distribution of classes on the data space \cite{mohammadi2022detection}. As the prototypes lie in the data space and encode typical behavior within classes, the output of GLVQ is interpretable. 
Some variants of GLVQ classifiers \cite{hammer2002generalized, schneider2009adaptive, bunte2012limited, saralajew2016adaptive, ghosh2020visualisation,tang2022generalized} use adaptive distances to improve the performance and to provide a deeper insight into the model's decision process. Generalized Relevance LVQ (GRLVQ)\cite{hammer2002generalized} introduced the relevance factors for features enabling applicants to observe which features play a bigger role in predictions. In this contribution, we extend the application of the GRLVQ to the Grassmann manifold and use relevance values to find relevant principal angles for image set classification (see Fig.\ \ref{fig:grassmann_mdf}).
The resulting method is intrinsically interpretable, both in terms of modelling classes, via prototype vectors, and finding important angles, by relevance factors. They provide a way to highlight the most influential images and pixels on predictions.

Having a model where its complexity depends on the size of data sets limits its application to small-size data sets. Several works, including  DCC, MSM, CMSM, PML, AMLS, use the Nearest Neighbor strategy for predictions, so they need to have access to all training examples. The kernel-based methods, i.e. GDA and GGDA, 
have a similar problem due to their need for computing pairwise kernel values between a new data point and all training examples. 
G(G)PLCR needs to store variation subspaces which means one variation subspace per training example. As a result, the complexity of the mentioned works makes them impractical for big data sets. 
As our proposed method is a member of GLVQ classifiers, it follows the Nearest Prototype strategy for prediction. 
Due to the fixed number of prototypes, determined independently of the dataset size, and typically much smaller than the dataset itself, there is a significant reduction in memory usage and time complexity during the testing phase. 

The major contributions of this paper contain:
\begin{itemize}
    \item Our method extends the application of the GRLVQ to the Grassmann manifold, such that prototypes become d-dimensional subspaces and inputs can be sets of vectors, which allows the learning and modeling of invariances. Similar to other LVQ models, its complexity does not depend on the dataset size and is pre-defined by the user.
    \item We apply it to several classification tasks, including handwritten digit recognition, face recognition, object recognition, and activity recognition. The experiments show that it outperforms previous works and can successfully model the possible variations, such as variations in handwritten styles and illumination conditions. Moreover, we demonstrate the role of relevance factors in removing redundant dimensions and reducing the effect of the manifold's dimensionality on the model's performance.
    \item We also showcase the model's transparency by offering detailed insights into its predictions. This is achieved through the highlighting of significant images and pixels that contribute to the model's decision-making process.
\end{itemize}

This paper is organized as follows: In section \ref{sec:background}, we provide a summary of the Grassmann manifold, as a way to represent data space, and the GLVQ classifier. In section \ref{sec:GLVQ4GM}, we first present the problem setting and then provide our proposed algorithm for dealing with it. Section \ref{sec:experiments} contains applications of the new method on five benchmarks, including two image classifications and three image-set classifications tasks. Finally, we conclude our work by highlighting some properties of the new technique.

\section{Background} \label{sec:background}
In this section, we provide the necessary details needed to explain our work. First, we give an overview of the Grassmann manifold representing the data space.
Then, we provide a short description of the Generalized Learning Vector Quantization (GLVQ) since we use it as a base for the new method.

\subsection{Grassmann manifold}\label{subsec:Grassmann}

A manifold is a topological structure that can be locally approximated as Euclidean space. A Riemannian manifold $\mathcal{M}$ is a manifold equipped with a distance, allowing to measure the distances 
on the manifold \cite{harandi2011graph}. 
\begin{definition}{(Geodesic distance)}
The geodesic distance $d_g$ between two points $p_1$ and $p_2$ on $\mathcal{M}$ is the length of the shortest path (on the manifold) connecting them.
\end{definition}

In this contribution, we focus on a specific type of Riemannian manifold, called the Grassmann manifold $\mathcal{G}$. 

\begin{definition}{(Grassmann Manifold)}
The Grassmann manifold $\mathcal{G} (D, d)$ is the set of all $d$-dimensional subspaces of $\mathbb{R}^D$ (where $d \leq D$).
\end{definition}
A popular way to model a point 
on $\mathcal{G} (D, d)$  is to represent it by a ($D \times d$)- orthonormal matrix $P$ such that its columns form a basis for its subspace. This representation is not unique since $span(P_1) = span(P_2)$ does not imply $P_1 = P_2$. 
In fact, it has been shown that  $\text{span}(P_1) = \text{span}(P_2)$ if and only if there exist $Q_1, Q_2 \in \mathcal{O}(d)$ (the group of $d \times d$ orthonormal matrices)  
such that $P_1 Q_1 = P_2 Q_2$. 

The Grassmann manifold $\mathcal{G} (D, d)$ inherits the geodesic distance from the Riemannian manifold. This distance is defined as the magnitude of the smallest rotation required to transform one subspace into the other \cite{harandi2013dictionary}. More precisely, it is a function of principal angles, which are recursively defined as follows:

\begin{definition}{(Principal angles/vectors)} \label{def:cc}
Let consider two subspaces $\mathcal{L}_1$ and $\mathcal{L}_2$ with the dimensionalities of $d_1$ and $d_2$, respecitvely. Then, $d$ \emph{principal angles}
\[
0 \leq \theta_1 \leq \theta_2 \leq \cdots \leq \theta_d \leq \frac{\pi}{2} \enspace, \hspace{0.4cm} d = \min{ \{d_1, d_2\} }
\]
are defined recursively by the following conditions:
  \begin{equation}
      \label{eq:canonicalcorr}
      \cos{\theta_i} = \max_{\substack{u_i^{\prime} \in \mathcal{L}_1 \\ v_i^{\prime} \in \mathcal{L}_2}} u^{\prime T}_i v^{\prime}_i = u_i^T v_i \enspace,
  \end{equation}
    subject to 
    \[
    \begin{cases}
    \lVert u_i^{\prime} \rVert = \lVert v_i^{\prime} \rVert = 1 \enspace,&\\
    u_i^{\prime T} u_j^{\prime} = v_i^{\prime  T} v_j^{\prime} = 0 \enspace, & \forall j < i \enspace,
    \end{cases}
    \]
   for all $i = 1, 2, \cdots, d$. Then, the cosine of principal angles is referred to as \emph{canonical correlation} and the elements of  $\{ u_k \}_{k=1}^d$ and $\{ v_k \}_{k=1}^d$ are known as \emph{principal 
   vectors}\cite{zhang2018grassmannian}.
\end{definition}

As can be seen, principal angles can be defined for any two subspaces. However, in this contribution, we assume that all subspaces are points on a Grassmann manifold, which implies that $d_1=d_2$. Consequently, sets of principal vectors, $\{ u_k \}_{k=1}^d$ and $\{ v_k \}_{k=1}^d$, form unitary bases for $\mathcal{L}_1$ and $\mathcal{L}_2$, respectively. Based on the principal angles, the geodesic distance on the Grassmann manifold is computed.
\begin{definition}{(Geodesic distance on the Grassmann manifold)}
Let $\mathcal{L}_1, ~\mathcal{L}_2 \in \mathcal{G} (D, d)$. Then, the geodesic distance between them is defined as:
\begin{equation}
    \label{eq:geo_dist}
    d_g(\mathcal{L}_1, \mathcal{L}_2) = \lVert \Theta \rVert_2 = \left( \sum_{i=1}^d \theta_i^2 \right)^{1/2}
\end{equation}
where $\Theta = [\theta_1, \cdots, \theta_d]$.
\end{definition}

Thus, to find distances among subspaces, we need principal angles.
In practice, the principal angles are calculated using the Singular Value Decomposition (SVD).
Let columns of $P_1$ and $P_2$ form unitary bases for two subspaces $\mathcal{L}_1$ and $\mathcal{L}_2$, respectively.
Principal angles can be found by the SVD of the matrix 
$M = P_1^T P_2 \in \mathbb{R}^{d \times d}$ \cite{bjorck1973numerical}, i.e.\ :
\begin{equation}
    \label{eq:compute_cc}
    P_1^T P_2 = Q_{12} (\cos{\Theta}) Q_{21}^T \enspace,
\end{equation}
where $\cos{\Theta} = \text{diag}(\cos{\theta}_1, \cdots, \cos{\theta}_d)$ and $Q_{12}Q^T_{12} = Q^T_{12}Q_{12} = Q_{21} Q_{21}^T = Q_{21}^T Q_{21} = I_d$. Thus,  canonical correlations are the singular values and their associated principal vectors are given by:
\begin{align}
	\label{eq:PQ}
	U &= P_1 Q_{12} = [u_1, \cdots, u_d] \enspace,\\
	V &= P_2 Q_{21} = [v_1, \cdots, v_d]\enspace,
\end{align}
where $Q_{12}$ and $Q_{21}$ can be seen as rotation matrices. 
In other words, 
we can obtain the principal vectors $U$ by rotating the orthonormal basis $P_1$ using the rotation matrix $Q_{12}$. Similarly, the principal vectors $V$ can be derived by rotating the orthonormal basis $P_2$ using the rotation matrix $Q_{21}$.

\subsection{Generalized Learning Vector Quantization (GLVQ)}\label{subsec:GLVQ}
A family of classifiers in the Explainable Artificial Intelligence (XAI) domain is Learning Vector Quantization (LVQ). 
Their ability to provide a clear explanation for their decisions makes them an appropriate tool for situations where we need insights into the decision reasoning.

Consider the training set $\{ (\vec{x}_i, y_i) \}_{i=1}^N$ where $\vec{x}_i \in \mathbb{R}^D$ and $y_i \in \{1, 2, \cdots, C\}$ represent a data point and its corresponding label, respectively. An LVQ model is represented by a set of labeled prototypes 
\[ 
\{ (\vec{w}_i, c(\vec{w}_i) \}_{i=1}^p \enspace,
\]
where $\vec{w}_i \in \mathbb{R}^D$ and $c(\vec{w}_i) \in \{1, \cdots, C\}$ denote a prototype vector and its corresponding label.
During the training, prototype vectors are distributed on the data space such that to represent the distribution of classes \cite{mohammadi2022detection}. 
For a prediction, it follows the Nearest Prototype strategy. It means 
for any new data point $\vec{x}$,  its label is predicted as the label of the closest prototype, i.e.:
\begin{align*}
     k &= \argmin _{i=1, \cdots, p} d( \vec{x}, \vec{w}_i ) \enspace,\\
     \tilde{c}(\vec{x}) &= c(\vec{w}_k) \enspace,
\end{align*}
where $d(\cdot, \cdot)$ denotes a measure to quantify the difference between vectors, and $\tilde{c}(\cdot)$ represents the model's prediction.
Therefore, to use the model for prediction, one only needs to save prototype vectors and their corresponding labels \footnote{Unlike the Nearest Neighbor strategy which needs to have access to all training examples.}.

In order to find the best position for prototypes, the Generalized Learning Vector Quantization (GLVQ) \cite{sato1995generalized} proposed the following cost function: 
\begin{align}
    E(\mathcal{W}) &= \sum_{i=1}^N \phi (\mu_i), 
    \text{with} ~ \mu_i = \frac{d(\vec{x}_i, \vec{w}_i^+) - d(\vec{x}_i, \vec{w}_i^-)}{d(\vec{x}_i, \vec{w}_i^+) + d(\vec{x}_i, \vec{w}_i^-)}\enspace,
    \label{eq:costfunction}
\end{align}
where $\mathcal{W}=\{ \vec{w}_1, \cdots, \vec{w}_p\}$ and $\phi$ is a differentiable monotonically increasing function. 
Here, we assume $\phi(x)=x$.
Moreover, $\vec{w}^+_i$ and $\vec{w}^-_i$ represent the closest prototypes to $\vec{x}_i$ with the same label and with a different label, respectively. 
From the cost function \eqref{eq:costfunction}, it is clear that the distance measure has a major role in the success of the GLVQ classifiers. Therefore, different variants of GLVQ classifiers, using different distances,  were proposed in 
 \cite{sato1995generalized, hammer2002generalized, schneider2009adaptive, bunte2012limited, ghosh2020visualisation}. 
For instance, GRLVQ \cite{hammer2002generalized} introduces an adaptive metric:
\begin{equation}
    d_{\lambda} (\vec{x}, \vec{w}) = \sum_{i=1}^D \lambda_i ( x^i - w^i )^2 ~~ \text{with} ~~ \lVert \lambda \rVert = 1 \enspace,
\end{equation}
where $\lambda_i \geq 0$ ($i=1, \cdots, D$), called relevance value, specifies the importance of $i$-th feature for separating classes.

To minimize the cost function, (stochastic) gradient descent learning has been used in \cite{hammer2002generalized}. For any randomly selected  sample $\vec{x}_i$, we only update the winning prototypes and the relevance factors:
\begin{align}
\vec{w}_i^{\pm} \leftarrow \vec{w}_i^{\pm} - \eta \frac{\partial E_s}{\partial \vec{w}_i^{\pm}} \enspace,\\
\lambda \leftarrow \lambda - \gamma \frac{\partial E_s}{\partial \lambda} \enspace,
    \label{eq:derE_w}
\end{align}
where $\eta$ and $\gamma$ are learning rates, and $E_s$ represents the cost value for $\vec{x}_i$.
During training, the model learns the optimal values for relevances $\{ \lambda_i \}_{i=1}^D$ and for prototypes $\{ \vec{w}_i \}_{i=1}^p$.

As explained in \cite{gilpin2018explaining}, an interpretable model must elucidate its inner workings in a manner understandable to users. The GRLVQ model attains this objective through the utilization of prototype vectors and relevance values. Since prototypes exist within the data space (i.e., $\mathbb{R}^D$), they can be inspected to extract characteristic values for each feature within a class. Additionally, the relevance values encode the significance of each feature in class differentiation, thereby accentuating the pronounced distinctions between classes. Consequently, the combined insight provided by prototypes and relevance values gives a rich understanding of classes and the model's underlying decision-making process.

\section{Generalized Learning Vector Quantization on the Grassmann Manifold} \label{sec:GLVQ4GM}
In this section, we first provide a problem description for image-set classification. Then, we present the new method.

\subsection{Problem setting} \label{subsec:problem_setting}
Let each image be represented as a $D$-dimensional vector. Then, the training set for an image-set classification task is:
\[
\{ (X_i, y_i) \}_{i=1}^N \enspace,
\]
where $X_i = [\vec{x}_{i1}, \cdots, \vec{x}_{im}] \in \mathbb{R}^{D \times m}$  denotes a data matrix of the i-th image set, containing $m$ images in its columns, and $y_i \in \{1, \cdots, C\}$ is its corresponding label.

As mentioned in section \ref{subsec:Grassmann}, an image set can be modelled as a linear subspace, 
generated by the columns of a unitary matrix. Hence, we use SVD to construct the matrix:
\begin{equation}
\label{eq:eigendecomposition}
X_i = P_i \Lambda R_i^T \enspace,
\end{equation}
where $P_i$ and $R_i$ contain left and right singular vectors, respectively, and $\Lambda$ is a diagonal matrix containing singular values in descending order. In practice, we only consider the first $d$ ($d \leq m$) columns of $P_i$ as a basis for the subspace. Thus, we represent the image set $X_i$ by $P_i \in \mathbb{R}^{D \times d}$. 

Following the idea of GLVQ classifiers, we initialize a set of prototypes that live on the data space, i.e. 
$\mathcal{G}(D, d)$\footnote{
Recently, Tang et al. \cite{tang2022generalized} proposed a novel approach which focuses on learning on the manifold of symmetric positive-definite matrices (SPD). However, this limits its applicability to situations where data is presented by SPD matrices. As a result, it is not suitable for the task of image set classification.
}:
\[
\{ (W_i, c(W_i)) \}_{i=1}^p \enspace,
\]
where $W_i \in \mathbb{R}^{D \times d}$ is an unitary matrix.
The cost function for the manifold can be re-written in a similar manner:
\[
E (\mathcal{W}) = \sum_{i=1}^N \phi ( \mu_i ) \enspace, \text{with} ~~ \mu_i = \frac{d_g (P_i, W_i^+) - d_g(P_i, W_i^-)}{d_g(P_i, W_i^+) + d_g (P_i, W_i^-)} \enspace,
\]
where $\mathcal{W} = \{W_1, W_2, \cdots, W_p\}$ and $d_g(\cdot, \cdot)$ is the geodesic distance on the Grassmanian manifold. 
Note that $W_i^-$ and $W_i^+$ are defined analogously to $w_i^-$ and $w_i^+$ as in Eq. \eqref{eq:costfunction}.

\subsection{Generalized Learning Grassmann Quantization}\label{subsec:GLGQ}
To find the best position of prototypes on the manifold, we need to optimize the cost function.
To achieve this, there are many variants of optimizers based on gradients. Here, we consider the stochastic gradient descent algorithm due to its low computational cost and its ability to handle big data sets. 
We compute the Euclidean gradient of the cost function. One may later use it to approximate the manifold gradient and apply more advanced techniques\cite{edelman1998geometry}.
 
Let $P$ be a randomly selected data point. We use the chain rule to compute the  gradient:
\begin{equation}
    \frac{\partial E_s}{\partial W^{\pm}} = \frac{\partial E_s}{\partial \mu} \frac{\partial \mu}{\partial d_g(P, W^{\pm})} \frac{\partial d_g(P, W^{\pm})}{\partial W^{\pm}} \enspace,
    \label{eq:derE_W}
\end{equation}
From the definition of $\phi$ and $\mu$, we obtain:
\begin{align*}
    \frac{\partial E_s}{\partial \mu} &= 1 \enspace,\\
    \frac{\partial \mu}{\partial d_g(P, W^{\pm})} &= \pm \frac{2  d_g(P, W^{\mp})}{\left( d_g(P, W^+) + d_g (P, W^-) \right)^2 } \enspace,
\end{align*}
To simplify the computation of the third term in Eq.\ \ref{eq:derE_W}, we use the square of the geodesic distance (i.e. $d_g \leftarrow d_g^2$):
\begin{equation}
    d_g(P, W) = \sum_{k=1}^d \theta_k^2 = \sum_{k=1}^d \arccos^2{u_k^T v_k} \enspace,
    \label{eq:rewrite-geodesic}
\end{equation}
where $u_k$ and $v_k$ are the principal vectors of subspaces generated by $P$ and $W$, respectively. 
From Eq.\ \ref{eq:rewrite-geodesic}, we get:
\[
\frac{\partial d_g(P, W)}{\partial v_k} = \frac{-2 \arccos{u_k^T v_k}}{\sqrt{1 - (u_k^T v_k)^2}} u_k \enspace,
\]
where $W$ can be $W^+$ or $W^-$.
In the matrix form, it is:
\begin{equation}
    \frac{\partial d_g(P, W)}{\partial V} = -UG \enspace,
    \label{eq:updaterule-geodesic}
\end{equation}
where $G$ is a diagonal matrix:
\[
[G]_{kk} = \frac{2 \arccos{u_k^T v_k}}{\sqrt{1 - (u_k^T v_k)^2}} \enspace,
\]
and $U=[u_1, \cdots, u_d]$, $V=[v_1, \cdots, v_d]$. 
We know that $U$ and $V$ are rotated versions of $P$ and $W$:
\begin{equation}
    U = PQ_P\enspace, \hspace{0.5cm} V = WQ_W\enspace,
    \label{eq:rotation_data_proto}
\end{equation}
where $Q_P, Q_W \in \mathcal{O}(d)$.
As a result, we have $\text{span}(W) = \text{span}(V)$, and they represent the same point on the $\mathcal{G}(D, d)$. 
Thus, the updating rule can be re-written as:
\begin{align}
    W^{\pm} &\leftarrow V^{\pm} - \eta \frac{\partial E_s}{\partial V^{\pm}} \enspace,
     \label{eq:modified_update_rule}
\end{align}
where 
\begin{align}
    \frac{\partial E_s}{\partial V^{\pm}} &= 
     \mp \frac{2  d_g(P_i, W^{\mp})}{\left( d_g(P_i, W^+) + d_g (P_i, W^-) \right)^2 } (U G)^{\pm} \enspace. 
    \label{eq:gradient_V}
\end{align}
Note that in general the Grassmann manifold is not closed under matrix subtraction. However, if prototypes are initialized as $d$ dimensional subspaces, for small learning rate $\eta$ the chance of having $W^{\pm}$ outside of $\mathcal{G}(D,d)$ (i.e. their dimensionalities become smaller than $d$) is negligible.

To utilize Eq.\ \eqref{eq:compute_cc} for calculating principal angles, it is necessary that the prototypes $W^{\pm}$ are represented as orthonormal matrices. To achieve this requirement, we use SVD to orthonormalize them after every update.
A summary of the algorithm can be found in Algorithm~\ref{alg:non_addaptive_alg}.

\RestyleAlgo{ruled}

\SetKwComment{Comment}{/* }{ */}

\begin{algorithm}[hbt!]
\caption{Generalized Learning Grassmann Quantization (GLGQ)}\label{alg:non_addaptive_alg}
\KwData{$\{ (P_i, y_i) \}_{i=1}^N$}
\KwResult{$\mathcal{W} = \{ (W_i, c(W_i)) \}_{i=1}^p$}
Randomly initialize prototypes with orthonormal matrices $\{W_1, \cdots, W_p\} $\;
$X \gets x$\;
\For{$i = 1, 2, \cdots, \text{nepochs}$}{
  Ids = Permute($1,2, \cdots, N$)\;
  \For{j in Ids}{
    \For{$k = 1, \cdots, p$}{
    $P_j^T W_k = Q_P \cos (\Theta_k) Q_{W_k}^T$\;
    Compute the geodesic distance Eq. \ref{eq:geo_dist}\;
    }
    Find prototypes $W^{\pm}$\;
    Compute gradients $\nabla_{V^{\pm}} E_s$, using Eq.\ \ref{eq:gradient_V}\;
    Update prototypes $W^{\pm}$, using Eq.\ \ref{eq:modified_update_rule}\;
    Orthonormalize the prototypes $W^{\pm}$\;
  }
}
\end{algorithm}
The proposed algorithm provides a set of labelled prototypes representing the typical behaviour within classes. These prototypes provide insight into decision-making logic by highlighting influential pixels for predictions. As a distance-based model, pixel roles are discerned through their impact on the computation of principal angles, and, as a result, on the geodesic distance. 
For a principal angle $\theta_i$ with a pair of principal vectors $u_i, v_i \in \mathbb{R}^D$, we have:
\begin{equation}
\cos{\theta_i} = u_i^T v_i = \sum_{j=1}^D u_{ij} v_{ij}\enspace.
\end{equation}
Hence, the impact of $k$-th pixel on $\theta_i$ is captured by the $k$-th element within the summation (i.e.~$u_{ik} v_{ik}$). 
This provides a way to see which pixels have bigger roles on $d_g$ and understand the internal decision-making process of the model.

\subsection{Generalized Relevance Learning Grassmann Quantization}\label{subsec:GRLGQ}

As Kim et al.\cite{kim2007boosted} have shown, each principal angle carries a different level of importance for separating 
classes. However, the geodesic distance $d_g$ assumes similar importance for all angles. 
Similar to \cite{kim2007boosted}, we propose an adaptive measure in which different principal angles can have different importance values. Then, we use data to learn their values. This has several advantages including giving a deeper insight into the decision process of the classifier by highlighting the most important angles. Moreover, it can reduce the sensitivity of the classifier in selecting the appropriate value for the dimensionality of subspaces $d$ since it automatically zeros the effect of redundant angles.

Motivated by Hammer et al.\cite{hammer2002generalized}, we define relevance factors of $\{ \lambda_k \}_{k=1}^d$ fulfilling the following condition: 
\begin{equation}
    \sum_{k=1}^d \lambda_k = 1, \quad \lambda_k \geq 0\enspace.
    \label{eq:relevance_condition}
\end{equation}
We introduce an adaptive distance measure:
\begin{equation}
    d_g^{\lambda} (P_1, P_2) = \sum_{k=1}^d \lambda_k \theta_k^2 = \sum_{k=1}^d \lambda_k \arccos^2{u_k^T v_k} \enspace.
    \label{eq:adaptive_dist}
\end{equation}
The relevance factors allow principal angles to have different levels of impact on measuring similarity between subspaces. 
To find both prototypes and relevance values, we use the stochastic gradient descent algorithm. 

Similar to the previous section, we derive:
\begin{equation}
    \frac{\partial E_s}{\partial V^{\pm}} = \mp \frac{2  d_g^{\lambda}(P, W^{\mp})}{\left( d_g^{\lambda}(P, W^+) + d_g^{\lambda} (P, W^-) \right)^2 } (UG)^{\pm}\enspace,
    \label{eq:update_rule_proto_adaptive}
\end{equation}
where $G$ is a diagonal matrix defined as:
\[
[G]_{kk} = \frac{2 \lambda_k \arccos{u_k^T v_k}}{\sqrt{1 - (u_k^T v_k)^2}} \enspace.
\]
Now, we need to find the derivative of the cost function with respect to the relevance factors. From the chain rule:
\begin{align}
    \frac{\partial E_s}{\partial \lambda_k} = \frac{\partial E_s}{\partial \mu_i} &\Big( \frac{\partial \mu_i}{\partial d_g^{\lambda}(P, W^{+})} \frac{\partial d_g^{\lambda}(P, W^{+})}{\partial \lambda_k} + \nonumber\\
    & \frac{\partial \mu_i}{\partial d_g^{\lambda}(P, W^{-})} \frac{\partial d_g^{\lambda}(P, W^{-})}{\partial \lambda_k} \Big)\enspace.
    \label{eq:grad_lamda}
\end{align}
From the definition of distance in Eq.\ \ref{eq:adaptive_dist}, we have:
\[
\frac{\partial d_g^{\lambda}(P, W)}{\partial \lambda_k} = \arccos^2{u_k^T v_k}\enspace.
\]
In the matrix form, it can be written as:
\[
\frac{\partial d_g^{\lambda}(P, W)}{\partial \lambda} = \text{diag}\Big( \arccos^2{(U^T V)}\Big)\enspace,  
\]
where diag(A) creates a diagonal matrix $B$ 
\[
[B_{ii}] = [A_{ii}]\enspace.
\]
Hence, the derivative for the relevance factors is as follows:
\begin{align*}
    \frac{\partial E_s}{\partial \lambda} &= \frac{2}{\Big( d_g^{\lambda} (P,W^+) + d_g^{\lambda} (P,W^-) \Big)^2} \cdot \biggl( \\  
    & d_g^{\lambda} (P, W^-) 
    ~\text{diag}\Big( \arccos^2{(U^{+T} V^+)}\Big)
    ~- \\
    &d_g^{\lambda} (P,W^+) ~
	\text{diag}\Big( \arccos^2{(U^{-T} V^-)}\Big)
     \biggl)\enspace.
\end{align*}
Then, updating rules will become:
\begin{align}
    \label{eq:updaterule_w} 
    W^{\pm} &\leftarrow 
    V^{\pm} - \eta \frac{\partial E_s}{\partial V^{\pm}}\enspace,\\
    \lambda &\leftarrow \lambda - \gamma \frac{\partial E_s}{\partial \lambda} \enspace.
    \label{eq:updatingrule_lamda}
\end{align}
Since the relevance factors are updated more often than prototypes, we set their learning rate $\gamma$ smaller (i.e., $\gamma \ll \eta$). 
Moreover, after every update, we orthonormalize the winning prototypes and normalize the relevance factors.
An overview of the algorithm is provided in Algorithm~\ref{alg:addaptive_alg}.

\RestyleAlgo{ruled}

\SetKwComment{Comment}{/* }{ */}

\begin{algorithm}[hbt!]
\caption{Generalized Relevance Learning Grassmann Quantization (GRLGQ)}\label{alg:addaptive_alg}
\KwData{$\{ (P_i, y_i) \}_{i=1}^N$}
\KwResult{$\mathcal{W} = \{ (W_i, c(W_i)) \}_{i=1}^p, \{\lambda_i\}_{i=1}^d$}
Randomly initialize prototypes with orthonormal matrices $\{W_1, \cdots, W_p\} $\;
Initialize the relevance values $\lambda = [\frac{1}{d}, \cdots, \frac{1}{d}] $\;
\For{$i = 1, 2, \cdots, \text{Nepochs}$}{
  Ids = Permute($1,2, \cdots, N$)\;
  \For{j in Ids}{
    \For{$k = 1, \cdots, p$}{
    $P_j^T W_k = Q_P \cos (\Theta_k) Q_{W_k}^T$\;
    Compute the distance (\ref{eq:adaptive_dist})\;
    }
    Find prototypes $W^{\pm}$\;
    Compute gradients $\nabla_{V^{\pm}} E_s, \nabla_{\lambda} E_s$ (\ref{eq:update_rule_proto_adaptive}, \ref{eq:grad_lamda})\;
    Update $W^{\pm}$ and $\lambda$ (\ref{eq:updaterule_w}, \ref{eq:updatingrule_lamda})\;
    Orthonormalize the prototypes $W^{\pm}$\;
    Normalize relevance factors $\forall l: \lambda_l \leftarrow \frac{\lambda_l}{\sum_t \lambda_t}$ \;
  }
}
\end{algorithm}

After training, the classifier provides not only prototypes but also a vector containing relevance factors. The relevance factors increase the transparency of the model since a user can inspect which principal angles/vectors have a significant role in separating classes, and as a result, in predictions. 
Knowing the most important principal vectors, one can inspect which images have the major role in their reconstruction. This is particularly useful when one inspects misclassifications and to determine which images caused the model failure. For a new data matrix $X$, we obtain from Eq.\ \ref{eq:eigendecomposition} and \ref{eq:rotation_data_proto}:
\begin{equation}
    U \approx X M\enspace,
    \label{eq:image_contribution}
\end{equation}
where $M=R \Lambda^{-1} Q_{P}$ captures the contribution of each image on principal vectors ($\Lambda^{-1}$ is a diagonal matrix with $[\Lambda^{-1}_{ii}] = [\frac{1}{\Lambda_{ii}}]$).
Additionally, relevance values reduce the effect of subspace dimensionality on the classifier's performance. 
We can pick a higher dimensionality, and then during the training process, the classifier diminishes the effect of unnecessary principal angles by decreasing (or zeroing) their corresponding relevance values.

\subsection{Complexity}
Representing a subspace by a unitary matrix is a common practice within methodologies developed for the Grassmann manifold. This is particularly the case for techniques that use principal angles to define distances, such as DCC, MSM, CMSM, BoMPA, DCC, MMD, GDA, GGDA, and AMLS(L).
Hence, SVD plays a central role across all these methods, facilitating the representation of subspaces and the computation of principal angles. Likewise, GRLGQ needs SVD to compute distances to prototypes, as well as to orthonormalize winning prototypes after each update. 
The computational complexity of SVD of a $(D \times d)$- matrix is $\textit{O}(D d^2)$. However, several strategies were proposed to reduce it to $\textit{O}(D d \log d)$\cite{golub2013matrix}. During training, GRLGQ needs to perform the SVD $N \times (p + 2) $ times per epoch. 
During test time, this reduces to $p$-times computations of singular values for a new subspace. 

During inference, the proposed method only requires access to $p$ prototypes. In contrast,  methodologies like DCC, GDA, GGDA, PML, AMLS(L) and G(G)PLCR necessitate access to all $N$ training examples due to their reliance on strategies such as Nearest Neighbor or kernel tricks.

\section{Experiments}\label{sec:experiments}
In this section, we apply the proposed method on several benchmark datasets to evaluate its performance. First, we consider two image classification tasks: 1) handwritten digit recognition, and 2) face recognition. Then, we assume three image-set classification tasks: 1) face recognition, 2) activity recognition, and 3) object recognition.

\subsection{Application on image classification task}
To apply GRLGQ to an image classification task, we first need to create a training set where data points represent $d$-dimensional subspaces. 
Here, we present a few steps to build such a set. Given a training set $\{ (\vec{x}_i, y_i) \}_{i=1}^n$, we perform the following steps for every class $c \in \{1, \cdots, C \}$:
\begin{enumerate}
\item randomly select $m$ images belonging to the class $c$,
    \[
    X_k = [\vec{x}_{k_1}, \vec{x}_{k_2}, \cdots, \vec{x}_{k_m}]
    \]
\item create the unitary matrix $P_k \in \mathbb{R}^{D \times d}$ (see Eq.\ \ref{eq:eigendecomposition}).
\end{enumerate}
At the end of this process, we obtain a new training set $\{ (P_k, c_k) \}_{k=1}^N$. To speed up the training process, one can initialize each prototype by the subspace constructed via applying PCA for each class.

In the test phase, new data points are images (1D subspace) while our prototypes are $d$-dimensional subspaces. Following the definition \ref{def:cc}, we can still compute the principal angle $\theta_1$ and use the following distance:
\[
d(\vec{x}, W) = \theta_1
\]
Thus, based on this measure, the model can assign a label to any vector $\vec{x}$ using the Nearest Prototype Strategy.

\subsubsection{Comparing GLVQ classifiers}
To compare the new method's performance with other GLVQ classifiers, we use the MNIST dataset, containing images of handwritten digits. 
It consists of 60k training images and 10k testing images where each image is a $28 \times 28$ grayscale. First, we converted images into vectors (of size 784) and then created a new training set consisting of subspaces (see above).

Among different GLVQ classifiers, we only considered GMLVQ\cite{schneider2009adaptive} and GTLVQ\cite{saralajew2016adaptive} due to their connections. 
Similar to our work, GMLVQ uses an adaptive distance for separating classes.
While our work uses subspaces to model variation within classes, the GTLVQ algorithm uses affine transformations to do the same.

Following \cite{saralajew2016adaptive}, we fixed the dimensionality of subspaces to $d = 12$, and we trained the model three times and reported the mean and standard deviation of the accuracies. 
We set the learning rates to $\eta = 10^{-4}, \gamma = 10^{-7}$, and the maximum number of epochs to 40.
The accuracies of classifiers (for one prototype per class) are shown in Table \ref{tab:LVQclassifiers}, where we compared our method to those reported in \cite{saralajew2020new}. It can be seen that GRGLQ outperforms others with the lowest number of parameters. 
A notable observation is that despite its enhanced flexibility by considering affine transformations, GTLVQ exhibits lower performance. This can be attributed to two plausible factors: a) due to its reliance on the Euclidean distance which tends to underperform in higher dimensions, and b) its presumption of equal importance levels across all base vectors.

\begin{table}
     \caption{The performance of different GLVQ classifiers on MNIST.}

     \vspace{-0.4cm}
   \centering
    \begin{tabular}{| c || c | c | c |}
         \hline
         Methods & Accuracy & $\#$ of parameters\\
         \hline
         GRLGQ & \textbf{94.78} $(\pm 0.03)$ & $784 \times 12 \times 10 + 12$ \\
         GTLVQ\cite{saralajew2020new} & $94.50 (\pm 0.05)$ & $784 \times ( 12 + 1 ) \times 10 \time 784$\\
         GMLVQ\cite{saralajew2020new} & $88.30 (\pm 0.05)$ & $784 \times 10 + 784 \times 784$\\
         \hline
    \end{tabular}
    \label{tab:LVQclassifiers}
\end{table}

\begin{figure}
    \centering
    \textbf{(a)}
    
    \includegraphics[scale=0.45]{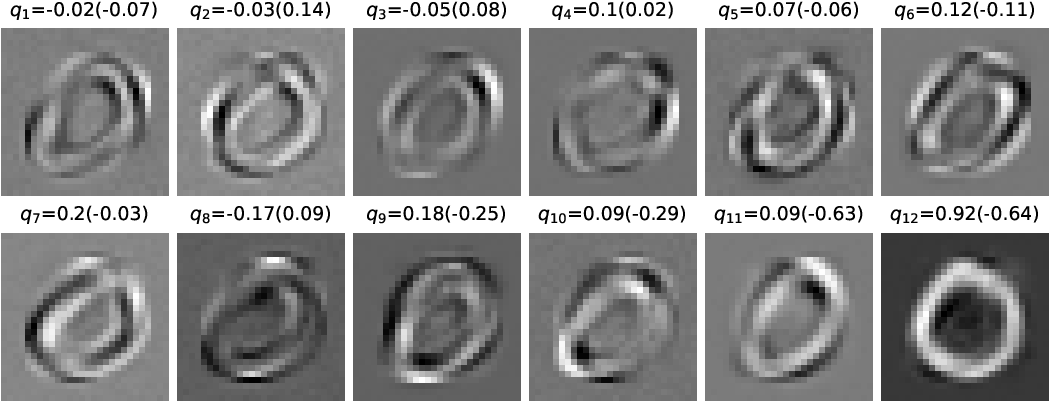}

    \textbf{(b)} \hspace{4cm} \textbf{(c)}
    
    \includegraphics[scale=0.6]{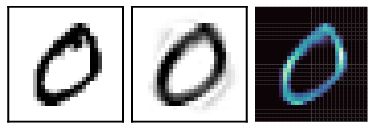}
    \hspace{0.5cm}
    \includegraphics[scale=0.6]{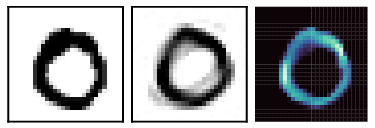}

    \caption{(a) displays principal vectors of the prototype for class 0. In (b) and (c), the left images show examples, the middle ones are the constructed images by the linear combination of principal vectors, and the left ones are heat maps of the contribution of pixels.
    Above prototype images, each number shows the coefficient of the corresponding principal vector on the linear combination for two examples in (b) and (c), respectively.
    }
    \label{fig:mnist_inspection}
\end{figure}

\begin{figure}
    \centering
    \hspace{1.2cm} \textbf{(a)} \hspace{2.9cm} \textbf{(b)}
    
    \includegraphics[scale=0.45]{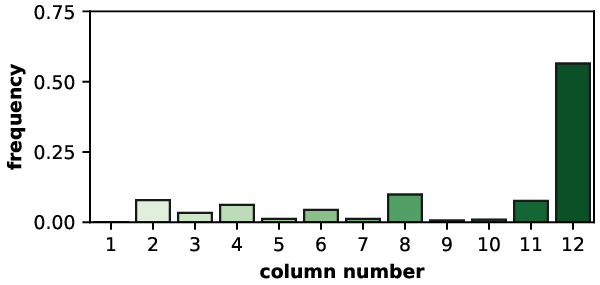}
    \hspace{0.1cm}
    \includegraphics[scale=0.45]{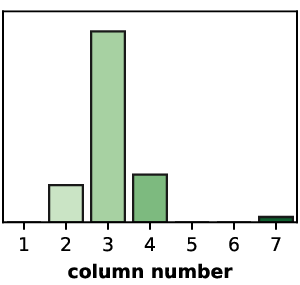}

    \caption{frequency of times each vector in the prototype matrix being the first principal vector with respect to training examples. a) MNIST, b) Yale Face database.}
    \label{fig:basline}
\end{figure}

Fig.\ \ref{fig:mnist_inspection} (a) displays the prototype of the class 0. It shows that principal vectors capture different handwritten styles.
In Fig.\ \ref{fig:mnist_inspection}(b-c), we show two different hand-writing with their corresponding images constructed from the prototype vectors. 
Let $\vec{x} \in \mathbb{R}^D$  ($||\vec{x}||=1$) be an image. To compute principal vectors, we need to perform SVD: 
\[
\vec{x}^T W = Q_{1 \times 1} \cos{\theta_1} (Q^W_{d \times 1})^T
\]
where subscripts show the dimensionality of matrices. Thus, the principal vector of the prototype is:
\[
V_{D \times 1} = W_{D \times d} Q^W_{d \times 1}
\]
This shows that the image $V$ is a linear combination of the columns of $W$, and the vector $Q^W$ contains their coefficient. Therefore,  $Q^W$ captures how columns of $W$ cooperate in constructing $V$ (see numbers above Fig.\ a). 
For example, it can be seen that the last image has the largest coefficient for the first example, and it defines a baseline for zero shapes. 
To formalize it, we create a plot for the number of times an image (in the prototype) has the highest effect for reconstructing training examples. 
The histogram is shown in Fig.\ \ref{fig:basline}a and demonstrates the capability of the last image in capturing the essence of the class 0. And the rest display different handwriting styles.
For instance, since the hand-written digit in Fig.\ \ref{fig:mnist_inspection}b is close to the baseline, the coefficient of the baseline has the highest effect $q_{12}  = 0.92$ (with a large margin). However, in Fig.\ \ref{fig:mnist_inspection}c, the role of the baseline reduces to $ q_{12} = -0.64$, and instead, the effect of the second image (which is more similar to the example) increases to $q_{2} =0.14$ in constructing the template image.

In addition to finding contributions of principal vectors, we can also find the most influential pixels (see section \ref{subsec:GLGQ}). In Fig.\ \ref{fig:mnist_inspection} (b-c) right,  heat maps are presented illustrating the influence of each pixel on the computation of the principal angle. This shows that pixels corresponding to handwritings play major roles in the prediction.
Consequently, the model effectively explains its decisions by reporting both the contributions of principal vectors and pixels.

\subsubsection{Comparison to other classifiers}
\begin{figure}
    \centering
    \includegraphics[scale=0.6]{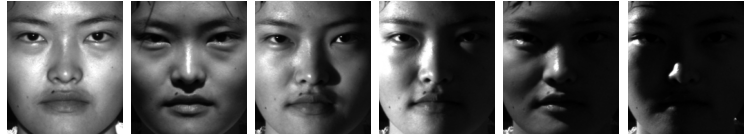}

    \caption{Examples of the Yale Face Database.}
    \label{fig:sample_YaleB}
\end{figure}

Here, we consider the Yale face Database \cite{deng2008four}. It consists of 2414 gray-level images of 38 people with different lighting conditions (see Fig.\ \ref{fig:sample_YaleB}). Following \cite{saralajew2016adaptive}, we resized the images (from $32 \times 32$) to $20 \times 20$ by bi-cubic spline interpolation, and we repeated a four-fold cross-validation process 10 times. 
Here, we set the learning rate to $\eta = 10^{-2}, \gamma = 10^{-5}$, and similar to \cite{saralajew2016adaptive},
we fixed $d=7$. As a stopping criterion, we considered the maximum number of 200 epochs.

We compared our method to those reported in \cite{saralajew2016adaptive, weinberger2009distance}, and Table~\ref{tab:otherclassifier} summarizes the results. 
GRLGQ provides the best performance with the lowest number of parameters.
While it only keeps 106407 values, GTLVQ should store 121600 parameters. This scenario worsens when considering kNN \cite{fix1989discriminatory} and LMNN \cite{weinberger2009distance}, as they require storage for the complete training set, thus incurring significant costs in terms of both memory usage and computational time.

\begin{table}
     \caption{Accuracies of different classifiers for the Yale Face Database.}

     \vspace{-0.3cm}
   \centering
    \begin{tabular}{| c || c | c | c |}
         \hline
         Methods & Accuracy \\
         \hline
         GRLGQ & \textbf{96.50} $(\pm 0.61)$\\
         GTLVQ\cite{saralajew2016adaptive} & $94.64 (\pm 0.68)$  \\
         SVM\cite{weinberger2009distance} & $84.80$ \\
         PCA with kNN\cite{weinberger2009distance} & $89.50$\\
         LDA with kNN\cite{weinberger2009distance} & $95.20$\\
         RCA with kNN\cite{weinberger2009distance} & $95.17$\\
         LMNN\cite{weinberger2009distance} & $94.10$\\
         multiple metric LMNN\cite{weinberger2009distance} & $95.90$\\
         \hline
    \end{tabular}
    \label{tab:otherclassifier}
\end{table}

\begin{figure}
    \centering
    \includegraphics[scale=0.45]{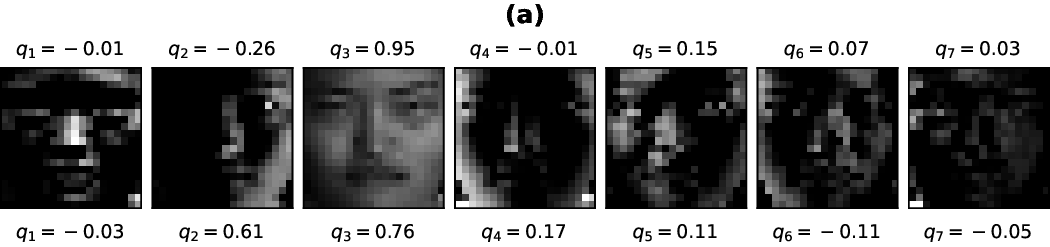}\\
    \vspace{0.2cm}
    \includegraphics[scale=0.5]{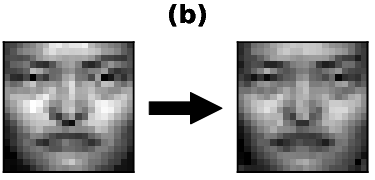}
    \hspace{0.7cm}
    \includegraphics[scale=0.5]{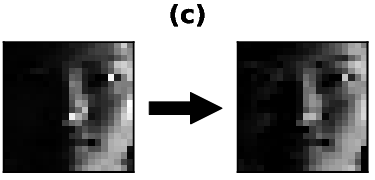}\\

    \caption{(a) displays principal vectors of the prototype for a class. In (b-c), the left images show two examples and the right ones are the constructed images by a linear combination of principal vectors.
    In the above and bottom of prototype images, numbers represent coefficients of the corresponding principal vectors on the linear combination for two examples in (b-c), respectively.}
    \label{fig:yaleface_inspection}
\end{figure}

In Fig.\ \ref{fig:yaleface_inspection}, we investigate prototypes and the way they contribute in constructing real images. From Fig.\ \ref{fig:yaleface_inspection}a, we see prototype vectors capturing different lighting conditions. Following the same process as the previous experiment, Fig.\ \ref{fig:basline}b shows that the third image of the prototype represents the baseline for the appearance of the face. Fig.\ \ref{fig:yaleface_inspection}b displays a face with good light; thus, the baseline has the highest contribution ($q_3 = 0.95$) on reconstructing the subject's face. However, in Fig.\ \ref{fig:yaleface_inspection}c, since the right side of the face is dark, the contribution of the second image increases to $q_2  = 0.61$ which causes the reconstructed image to resemble the real image (see Fig.\ \ref{fig:yaleface_inspection}c right). 

\subsection{Application on image-set classification task}
In the subsequent sections, we consider image-set classification tasks across three benchmark datasets. 
Due to the unavailability of source code for some of the existing works, we have decided to compare our method's performance against the results reported in two works \cite{wei2020prototype,sogi2020metric} that exhibit higher accuracy. To ensure a fair comparison, we replicated the preprocessing pipeline outlined in \cite{wei2020prototype,sogi2020metric}.
After preprocessing, for every image set, we transform it into an orthonormal matrix through the following steps:
\begin{enumerate}
\item store images as columns inside a matrix, say $X_i$,
\item create the unitary matrix $P_i \in \mathbb{R}^{D \times d}$ (see Eq.\ \ref{eq:eigendecomposition}),
\end{enumerate}
This procedure creates a training/test set, denoted as $\{(P_i, y_i)\}_{i=1}^N$, wherein $P_i$ represents the subspace generated by the $i$-th image-set.

In the following experiments, the learning rates are fixed to $\eta=0.05, \gamma=0.0001$. Moreover, to speed up the convergence of training, we initialized prototypes with randomly selected examples for each class.

\subsubsection{Face recognition}
The Extended Yale Face Database B (YaleB) contains 16,128 images of 28 people. For each person, images are taken under different poses and illumination conditions (9 poses and 64 illumination conditions). 
To compare our algorithm with the results presented in \cite{wei2020prototype}, we followed the same preprocessing step. This involves the following steps: a) we applied the cascaded face detector to extract faces and resized them to $20 \times 20$. b) 
for each person, we grouped images by their pose, resulting nine image sets for each subject. 
c) for each person, we randomly selected three image sets for training, and the rest were used for testing. Note that similar to \cite{wei2020prototype}, we repeated (c) ten times and then computed the mean and standard deviation of accuracies.

\begin{figure}
    \centering
    \includegraphics[scale = 0.55]{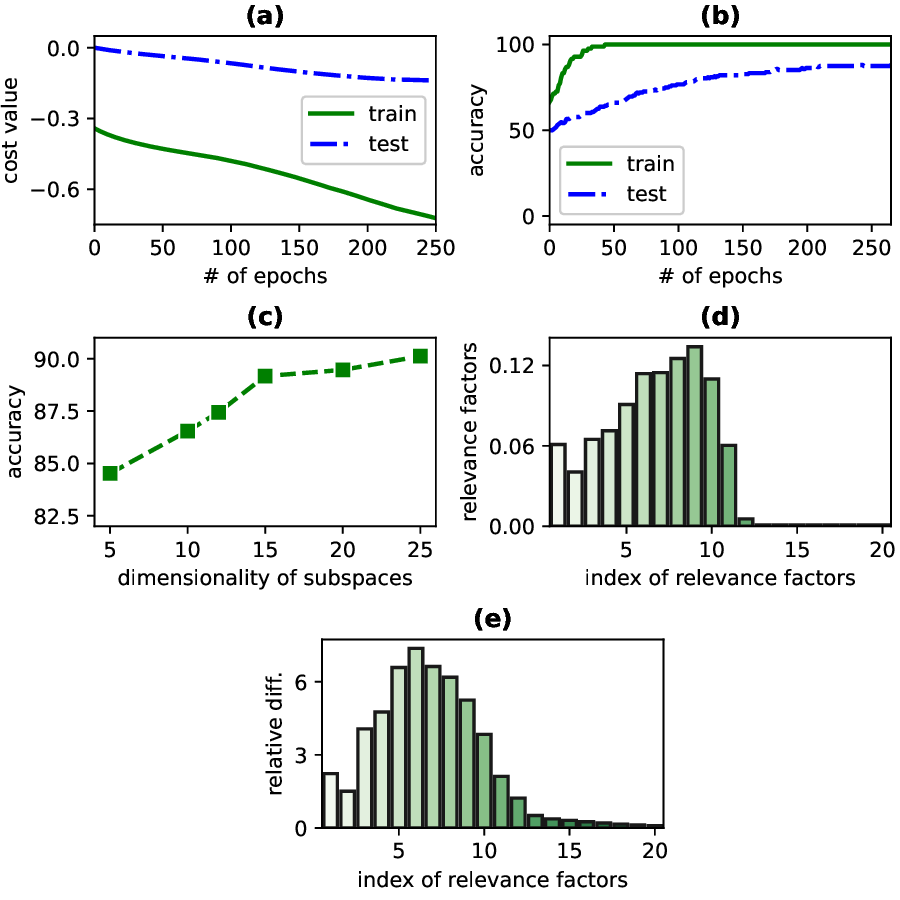}

    \caption{(a) and (b) display the cost function values and accuracies during training. The performance of our algorithm for different dimensionalities is given in (c);  (d) shows the learned relevance factors; i-th column in (e) encodes the mean of differences between square of i-th principal angles of $W^{\pm}$ (i.e. $\frac{(\theta_i^-)^2 - (\theta_i^+)^2}{(\theta_i^+)^2}$).}
    \label{fig:cost_rel-ExYaleB}
\end{figure}

Fig.\ \ref{fig:cost_rel-ExYaleB} (a-b) demonstrates how the optimization process can successfully decrease (increase) the cost function (accuracy) values over training. Moreover, Fig.\ \ref{fig:cost_rel-ExYaleB}c displays the accuracy of the model for different dimensionalities $d$. It shows that the performance does not reduce when we increase the dimensionality. In fact, it even improves since the model sees more details and can decide (using data) what to pick for better performance.
This does not extend to earlier methods. As outlined in \cite{wei2020prototype}, the choice of dimensionality has a high impact on their performance.

Table \ref{tab:experiments-exYaleB} summarizes results and demonstrates that our method outperforms other methods on this task. (G)GPLCR \cite{wei2020prototype} provides the second-best performance. 
To compare the models' complexities, we have computed their respective parameter counts as follows:
\begin{align*}
\text{G(G)PLCR}&: 
\big( \text{\# of classes} + \text{size of training set} \big) \times D \times d \\
 &= (28 + 3 \times 28) \times 400 \times 3 = 134400\\
 \text{GRLGQ}&: \big( \text{\# of classes} \big) \times D \times d + d  \\
 &=28 \times 400 \times 25 + 25 = 280000
\end{align*}
This demonstrates that the parameter count for GRLGQ is nearly twice that of G(G)LPCR. Furthermore, the parameter count for G(G)LPCR depends on dataset size and does not scale up for big data sets. In contrast, GRLGQ provides good performance regardless of the dataset size, as its complexity does not depend on the number of data points.

\begin{table}
     \caption{Accuracies of classifiers for the Ex. YaleB and ETH-80.}

     \vspace{-0.3cm}
   \centering
    \begin{tabular}{| c || c | c |}
         \hline
         Methods & Ex. Yale B & ETH-80 \\
         \hline
         DCC & $77.08 (\pm 3.25)$ & $90.75 (\pm 3.34)$ \\
         MSM & $70.00 (\pm 4.12)$ & $77.00 (\pm 6.21)$ \\
         AHISD & $55.95 (\pm 5.54) $& $73.00 (\pm 7.24)$ \\
         CHISD &$ 62.38 (\pm 4.97)$& $75.50 (\pm 4.38)$ \\
         GDA \cite{wei2020prototype}& $82.80 (\pm 0.96)$ & $91.00 (\pm 2.13)$\\
         GGDA \cite{wei2020prototype}& $79.11 (\pm 0.92)$ & $92.50 (\pm 1.16)$\\
         PML \cite{wei2020prototype} & $77.08 (\pm 0.18)$ & $93.75 (\pm 1.80)$\\
         AHISD \cite{wei2020prototype}& $59.05 (\pm 0.89)$ & $70.00 (\pm 0.68)$\\
         GPLCR \cite{wei2020prototype}& $88.45 (\pm 0.56)$ & $95.25 (\pm 1.24)$\\
         GGPLCR \cite{wei2020prototype}& $ 88.57 (\pm 0.48)$ & \textbf{96.75 $(\pm 1.30)$}\\
         GRLGQ & \textbf{90.12 $(\pm 2.25)$} & $94.5 (\pm 1.87)$\\ 
         \hline
    \end{tabular}
    \label{tab:experiments-exYaleB}
\end{table}

\begin{figure}
    \centering
    \textbf{(a)}
    
    \includegraphics[scale=0.55]{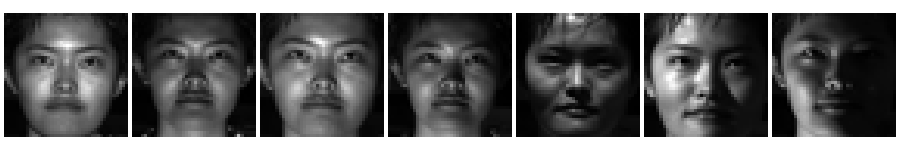}
    
    \textbf{(b)}
    
    \includegraphics[scale=0.53]{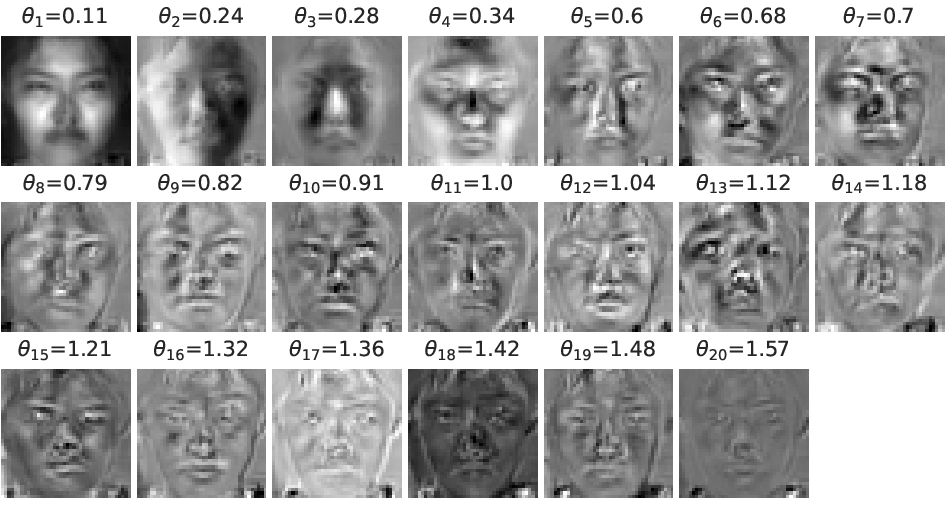}
    
    \textbf{(c)}
    
    \includegraphics[scale=0.53]{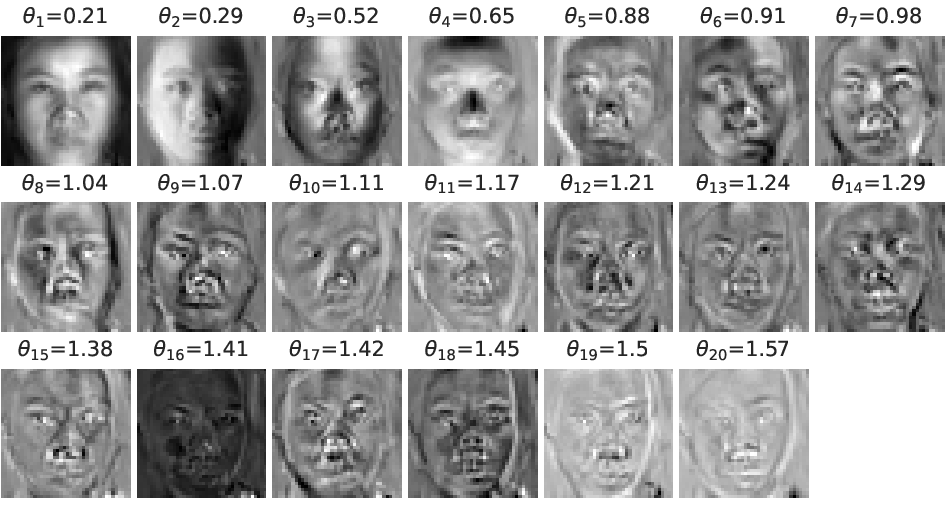}

    \caption{a) a few images of a person under different light conditions. b) the closest prototype with the same label (i.e. $W^+$). c) the closest prototype with a different label (i.e. $W^-$).}
    
    \label{fig:img_ExyaleB}
\end{figure}

To examine the output of the algorithm visually, we applied our algorithm to images with higher resolution ($40 \times 40$).
Fig. \ref{fig:img_ExyaleB} showcases a few shots of an image set alongside its corresponding winning prototypes (i.e. $W^{\pm}$). 
If one compares principal angles of $W^+$ to the same one in $W^-$, the difference between them for the angles with small relevance values (see Fig.\ \ref{fig:cost_rel-ExYaleB}d) is relatively small. Fig.\ \ref{fig:cost_rel-ExYaleB}e also confirms it since it displays the mean of relative differences between (the square of) principal angles of winning prototypes. It can be seen that it resembles the relevance values profile. 
Therefore, \ref{fig:cost_rel-ExYaleB}(d-e) represent that relevance factors can successfully diminish the effect of the redundant principal angles. 
This explains why increasing the dimensionality of subspaces does not have a negative effect on the performance of the algorithm. In addition to zeroing the effect of redundant principal angles, Fig.\ \ref{fig:cost_rel-ExYaleB}d shows the algorithm also gives small weights to the first few angles, which can be explained by seeing the same behavior in Fig.\ \ref{fig:cost_rel-ExYaleB}e. 
Since their values are close (in comparison to angles with bigger weights), their capability in separating classes is lower. 

\subsubsection{Object recognition}
For the object recognition task, we use the ETH-80 \cite{leibe2003analyzing} which contains 8 categories of objects, including apples, cars, cows, cups, dogs, horses, pears, and tomatoes. 
For each category, there are ten object subcategories consisting of 41 images (taken from different viewpoints). Here, we considered each subcategory as an image set. 
Following previous works \cite{sogi2020metric, wei2020prototype}, as a pre-processing step, we first converted them into grayscale images and then resized them to $20 \times 20$. To create a training and testing set, we randomly selected five subcategories of each object for training and the remaining five for testing. We repeated it 10 times and then reported the averages and standard deviations of accuracies. Here, we set the dimensionality $d$ to $5$ for the GRLGQ algorithm, and the result is reported in Table \ref{tab:experiments-exYaleB}.

Although G(G)PLCR delivers superior performance, it comes at the cost of memory complexity tied to data size. As a result, it stores 60000 parameters. In contrast, parameter count for GRLGQ, the third best method, does not depend on the data volume, so it is one-fourth of that of G(G)PLCR (i.e., 16,005).

\begin{figure}
    \centering
    \textbf{(a)}

    \includegraphics[scale=0.85]{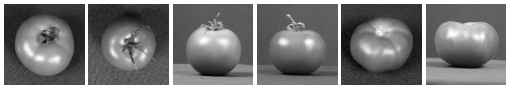}
    \includegraphics[scale=0.85]{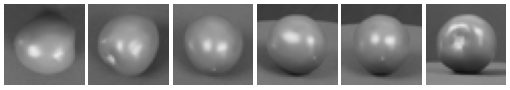}

    \textbf{(b)}
    
    \includegraphics[scale=0.5]{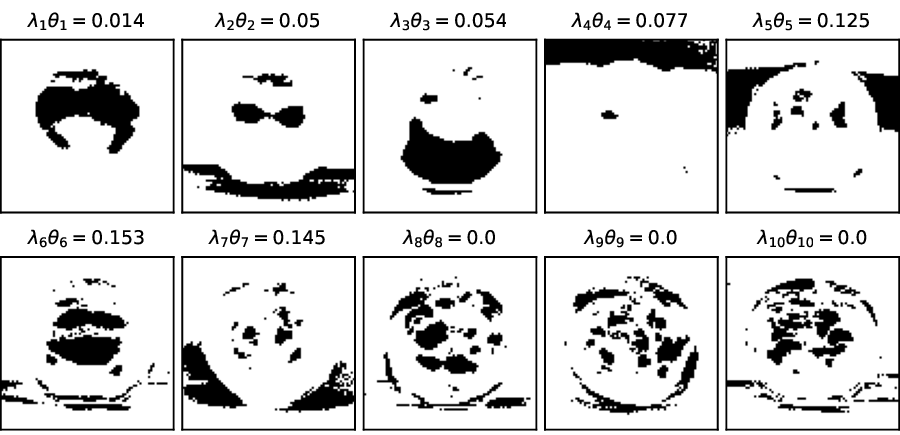}
    
    \textbf{(c)}
    
    \includegraphics[scale=0.5]{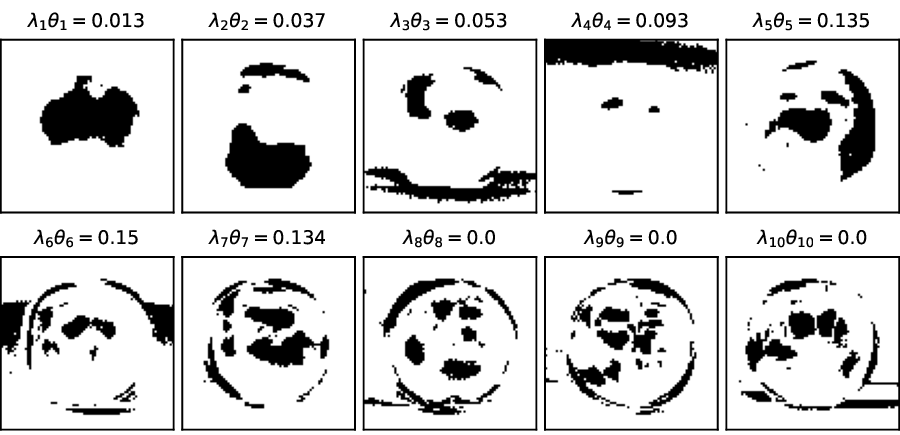}

    \textbf{(d) 2nd}  \hspace{1.8cm} \textbf{(e) 7th}

    \includegraphics[scale=0.5]{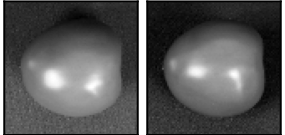} 
    \hspace{0.5cm}
    \includegraphics[scale=0.5]{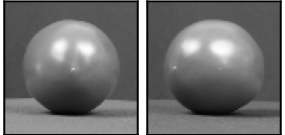}
    
    \caption{a) The first row displays different subcategories of tomatoes and the second row shows images of a  misclassified tomato. For the misclassified tomato, b and c depict 15\% of pixels of prototypes (for tomato and apple) with the highest response. The above numbers show the contribution of principal vectors on the distance. (d) and (e) depict two images with the highest contribution to reconstructing two principal vectors with the highest impact on the misclassification.}
    \label{fig:pixels_eth80}
\end{figure}

In Figure Fig.\ \ref{fig:pixels_eth80}a, tomato subcategories are presented in the first row, accompanied by a few images of a misclassified tomato (classified as an apple) in the second row.
For a closer examination of the model's rationale behind misclassifications, Fig.\ \ref{fig:pixels_eth80}(b-c) display 15\% of pixels with the highest impact on the computation of principal angles, as discussed in section \ref{subsec:GLGQ}. Through a comparison of the principal angles' influence on the distance computation (numbers above images) for prototypes of both tomato and apple, it becomes evident that the most pronounced disparities, causing misclassification, arise in the second, and seventh principal vectors. Interestingly, they share a common attribute: they emphasize distinct aspects. Specifically, while the pivotal pixels for the tomato concentrate on the background, the decisive pixels for the apple include those corresponding to the tomato itself. 
This might stem from the tomato's apple-like shape, compounded by the absence of a sepal-like structure.  This is confirmed by Fig.\ \ref{fig:pixels_eth80}(d-e) depicting images with the highest contribution on the two principal vectors (see Eq.\ \ref{eq:rotation_data_proto}).  
This leads to a slightly smaller distance for the apple's prototype ($0.465 < 0.468$).

\begin{figure} 
	\centering
	\textbf{(a)} \hspace{4cm} \textbf{(b)}
	
    \includegraphics[scale=0.5]{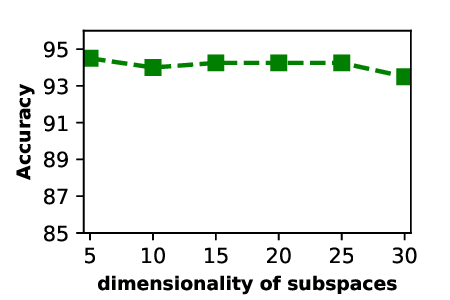}
    \hspace{0.1cm}
    \includegraphics[scale=0.5]{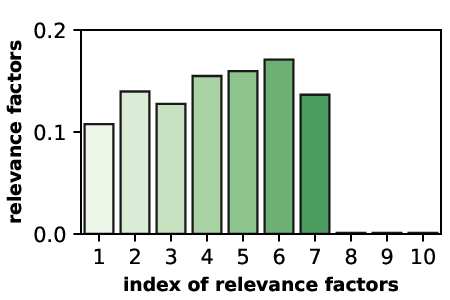}
    
    \textbf{(c)}
    
    \includegraphics[scale=0.4]{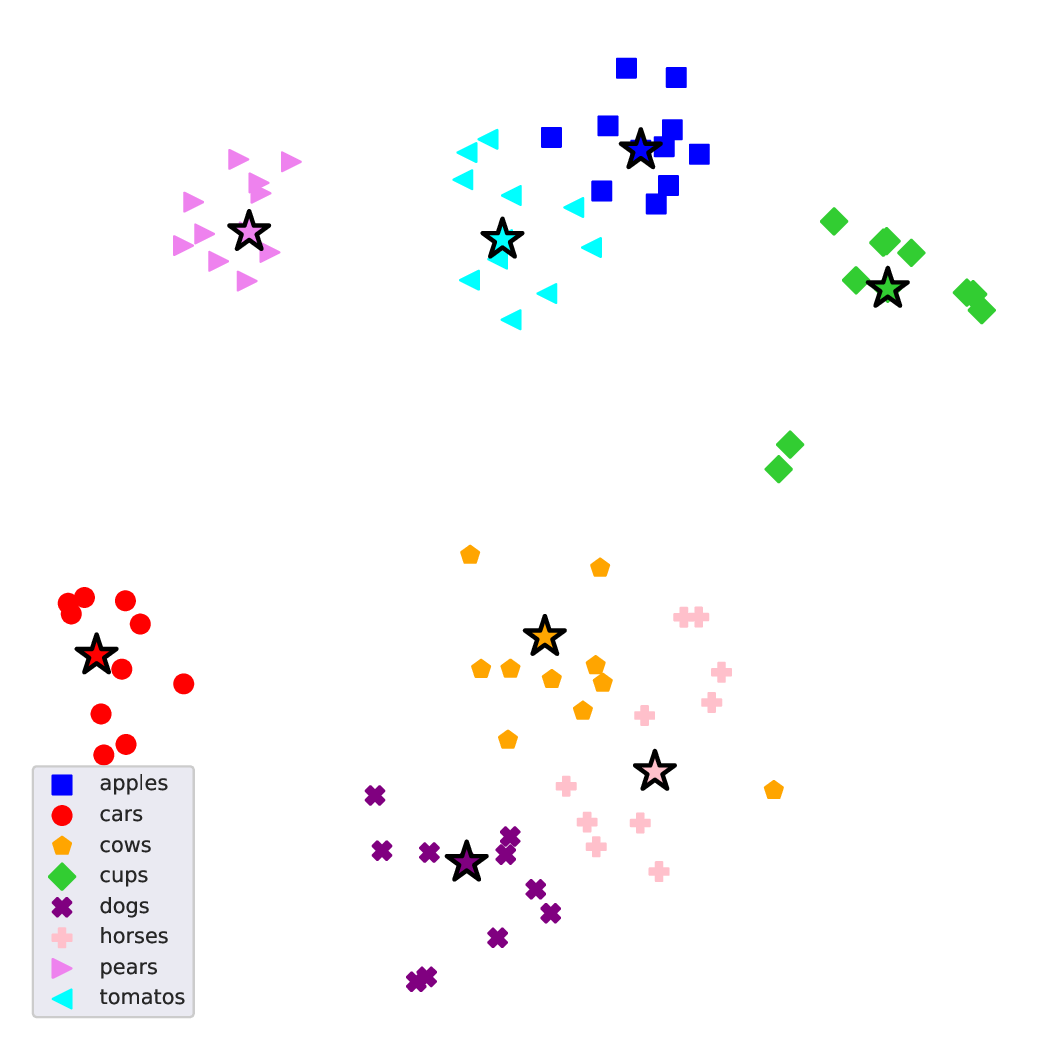}

    \caption{a) The performance of GRLGQ for different dimensionalities. b) The relevance profile. c) Visualization of the data set and prototypes (stars) using t-SNE.}
    \label{fig:tsne_eth80}
\end{figure}

As it has been shown in \cite{wei2020prototype} regarding the ETH-80 data set, the choice of the subspace dimensionality $d$ exerts significant influence over the performance of methodologies developed on the Grassmann manifold. Specifically, their accuracy drops to below 90\% when $d$ exceeds 3.
However, as illustrated in Fig.\ \ref{fig:tsne_eth80}a, GRLGQ stands apart by demonstrating remarkable stability across a wide range of dimensionalities. In essence, the efficacy of relevance factors becomes evident in mitigating the influence of extraneous principal angles. This is exemplified by the visualization in Fig.\ \ref{fig:tsne_eth80}b, wherein the redundancy of the last three principal angles becomes apparent for $d=10$, with relevance values effectively zeroing their impact.

Finally, to visualize the spatial arrangement of prototypes within the dataset, we employ t-distributed Stochastic Neighbor Embedding (t-SNE) \cite{van2008visualizing} applied to the distance matrix. This matrix contains pairwise distances among data points and also prototypes. As depicted in Fig.\ \ref{fig:tsne_eth80}c, the prototypes are positioned at the center of their respective classes. This placement not only reduces the distance between class members and their corresponding prototypes but also amplifies the separation from other classes. This observation serves as evidence that our optimization algorithm has effectively identified optimal prototypes and relevance factors.

\subsubsection{Activity recognition}
UCF Sports is a dataset containing videos of 10 actions including 
diving, 
golf swing, 
kicking, 
lifting, 
riding horse, 
running, 
skateboarding, 
swing-bench, 
swing-side, 
walking. 
It consists of 150 videos (containing between 22 to 64 frames) with a resolution of $720 \times 480$. The actions happened in very different scenes and were captured from different viewpoints \cite{rodriguez2008action}. To compare our result with those in \cite{sogi2020metric}, we followed the same preprocessing steps as follows: We first extracted bounding boxes containing humans (provided with the dataset). Then, we converted images to grayscale level and resized them to $38 \times 24$. 
Similar to \cite{rodriguez2008action,sogi2020metric}, we used a leave-one-out cross-validation scheme (LOOCV) to see the performance of different algorithms. Thus, the training process was repeated 150 times (149 videos for training and 1 video for testing). 
As shown in previous experiments, GRLGQ's performance is stable over a wide range of values. Thus, we set $d=22$ which is the minimum number of frames for a video. This allows the model to automatically zero unnecessary principal vectors/angles.
The results are shown in Table \ref{tab:experiments-ucf}.

\begin{table}
     \caption{Accuracy for the UCF data set (other classifiers' results from \protect\cite{sato1995generalized}).}
     
     \vspace{-0.3cm}
   \centering
    \begin{tabular}{| c || c |}
         \hline
         Methods & UCF  \\
         \hline
         DCC & 59.33 \\
         MSM & 48.67\\
         AHISD & 53.33 \\
         CHISD & 52.00\\
         GDA \cite{sogi2020metric}& 70.00 \\
         PML \cite{sogi2020metric}& 72.67 \\
         AMLS \cite{sogi2020metric}& 71.33 \\
         AMLSL \cite{sogi2020metric} & 74.00 \\
         GRLGQ & \textbf{80.00} \\
         \hline
    \end{tabular}
    \label{tab:experiments-ucf}
\end{table}

It can be seen that our method achieves the best performance with a large margin. The next method is AMLSL \cite{sogi2020metric}. Similar to our work, it uses an adaptive distance, which is learned during training. However, while our method uses the Nearest Prototype strategy for prediction, AMLSL follows the Nearest Neighbor strategy which means it needs to have access to the training set for prediction. Therefore, since the number of prototypes is (much) smaller than the size of the test set, our method provides an efficient solution, in terms of lower time and memory during the test phase. 

\begin{figure}
    \centering
    \includegraphics[scale=0.5]{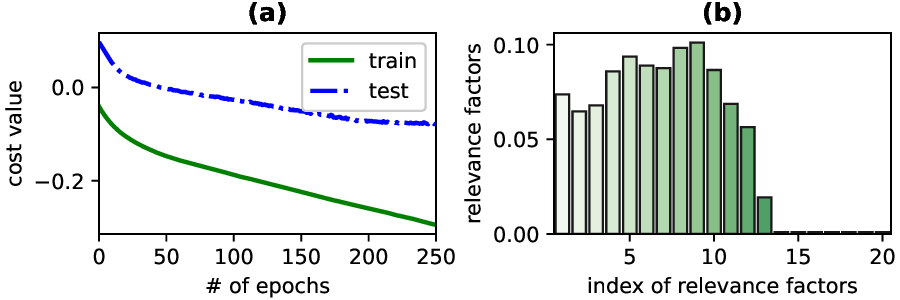}
    
    \footnotesize\textbf{(c)}
    
    \includegraphics[scale=0.38]{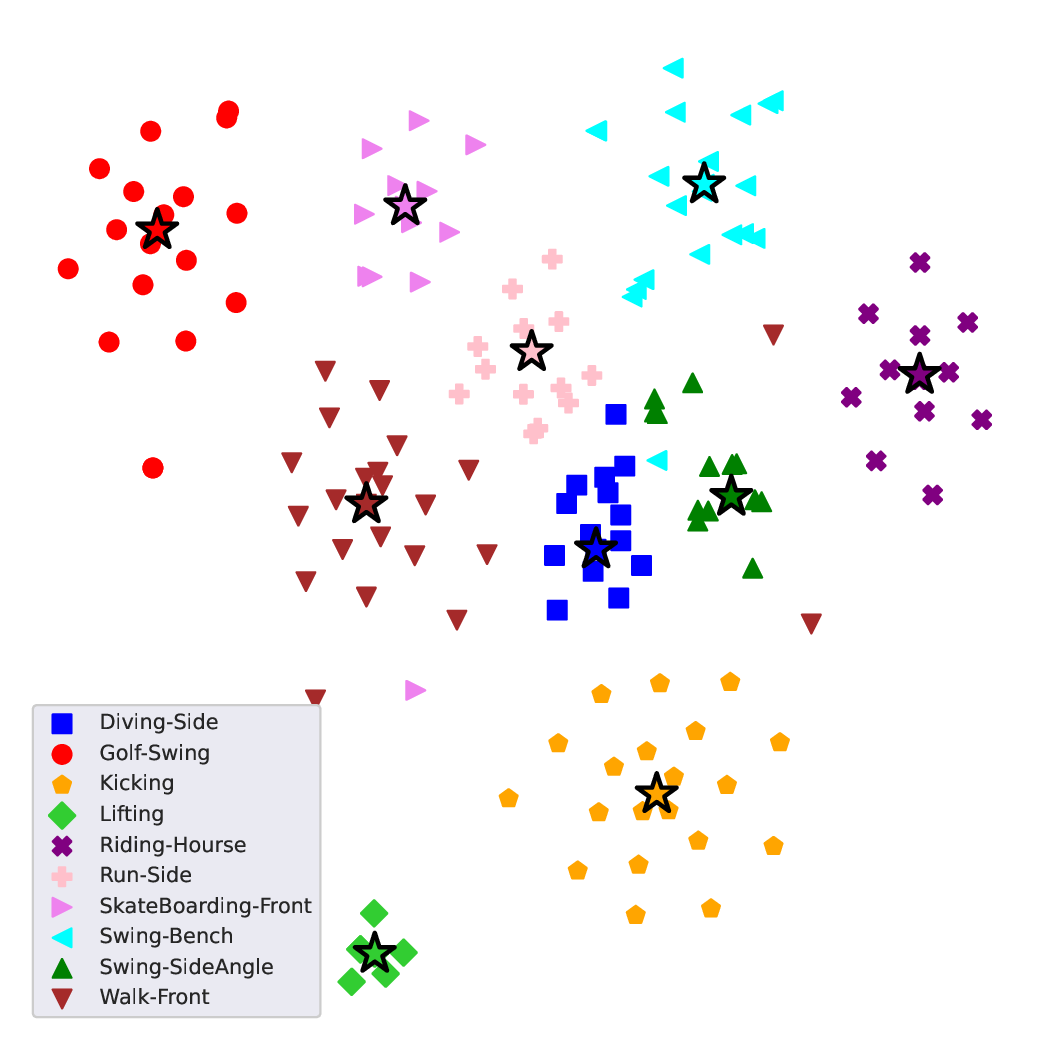}

    \caption{a) cost function values along different epochs. b) the learned relevance factors.
    c) visualization of data through the application of t-SNE on the distance matrix (prototypes are represented by stars).}
    \label{fig:cost_rel_ucf}
\end{figure}

\begin{figure}
    \centering
    \text{(a)}
    
    \includegraphics[scale=0.83]{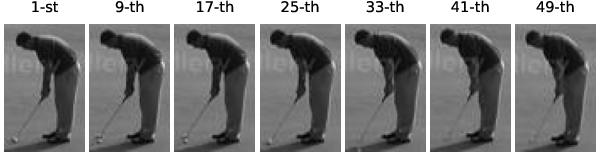}
    
    
    
    \text{(b)}
    
    \includegraphics[scale=0.7]{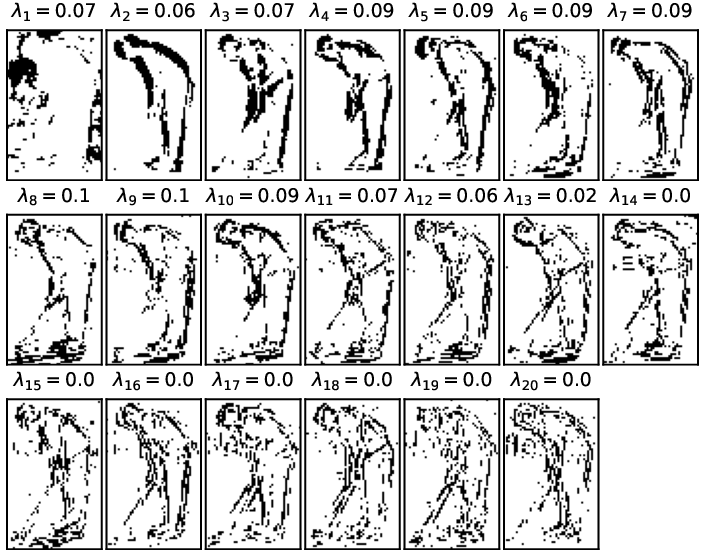}

    \caption{a) a few frames of an image set for the 'golf-swing' class.
    b) the prototype (with 20 principal vectors) for the 'golf-swing' class, with 15\%  of pixels highlighted to denote their significant impact on prediction. The value $\lambda$ shown on top of each image represents its corresponding relevance factor. Notably, the prototype gives the highest importance to pixels outlining the person, particularly for the vectors with bigger relevance value.}
    \label{fig:proto_plus_ucf}
\end{figure}

To visually inspect GRLGQ's output, we applied the algorithm on double-sized images (i.e. $76 \times 48$) with $d=20$. Fig.\ \ref{fig:cost_rel_ucf} (a-b) show changes in cost values over training and the resulting relevance profile, respectively. As seen, the model found the last six principal angles irrelevant for the classification task and zeros their effect on the computation of distances. 
To get a deeper understanding, we utilized the model to predict the label of the image set depicted in Fig.\ \ref{fig:proto_plus_ucf}a. 
In Fig.\ \ref{fig:proto_plus_ucf}b, we highlight 15\% of pixels with the highest impact on each principal angle. Evidently, these sketch the shape of the player's body when he hits the ball. Consequently, the model identifies the player's body state during the ball hit as the most discriminative attribute for differentiation of this activity from others.

In order to gain insights into the distribution of prototypes across the data space, we employed the t-SNE algorithm to visualize both data points and prototypes. As depicted in Fig.\ \ref{fig:cost_rel_ucf}c, similar to the earlier experiment, the prototypes are positioned proximate to the centre of each class.

\section{Conclusion} \label{sec:conclusion}

Using multiple images for prediction, instead of a single image, helps to create more robust classifiers.
Therefore, several techniques \cite{yamaguchi1998face, fukui2005face, kim2007boosted,  kim2007discriminative, wang2008manifold, wang2009manifold, hamm2008grassmann, harandi2011graph, huang2015projection, wei2020prototype, sogi2020metric} have been developed in which image-sets are represented as subspaces. These subspaces form a non-linear manifold, called the Grassmann manifold, and previous methods have tried to perform the classification on it.

In this contribution, we present the Generalized Relevance Learning Grassmann Quantization (GRLGQ), which is applicable for the classification on the Grassmann manifold. 
The new method returns a set of prototypes $\{ (W_i, c(W_i) \}_{i=1}^p$ and a set of relevance factors $\{ \lambda_i \}_{i=1}^d$.
Both sets provide important details about the decision-making process of the model. 
Prototypes (a set of principal vectors) serve as representatives of individual classes, illuminating the pixels pivotal for class differentiation. 
Meanwhile, relevance factors introduce an adaptive distance metric, endowing prototype elements with varying levels of importance. 
This establishes an intrinsically interpretable model, meaning it provides insightful details about its predictions. Thus, a user can see how meaningful is the decision of the model during inference time.

Through experiments, we have demonstrated the superiority of the proposed method over previous works on the image-set classification task. In contrast to earlier methodologies, the complexity of the model can be set by the user and remains uninfluenced by the scale of the training set. Additionally, the learned distance measure within the GRLGQ framework contributes to the development of a model that exhibits reduced susceptibility to dimensionality choices, thus yielding consistent outcomes across various dimensionalities – a characteristic not found in prior works.
In summary, while the proposed model provides better performance, it keeps the complexity under control and provides an explanation of its predictions.

\bibliographystyle{IEEEtranIES}
\bibliography{mybibfile}

\EOD

\end{document}